\documentclass{article}
\usepackage{ijcai21}

\usepackage{times}
\usepackage{soul}
\usepackage{url}
\usepackage[hidelinks]{hyperref}
\usepackage[utf8]{inputenc}
\usepackage[small]{caption}
\usepackage{graphicx}
\usepackage{amsmath}
\usepackage{amsthm}
\usepackage{booktabs}
\usepackage{algorithm}
\usepackage{subfig}
\usepackage{algorithmic}
\usepackage[normalem]{ulem}
\useunder{\uline}{\ul}{}
\title{Motion Detection using CSI from Raspberry Pi 4} 

\author{
    Glenn Forbes$^1$
    \and
    Stewart Massie$^1$
    \and 
    Susan Craw$^1$
    \and
    Christopher Clare$^2$
    
    \affiliations
    $^1$ Robert Gordon University, Aberdeen, UK
    $^2$ Antillion Ltd, Bristol, UK
    \emails
    \{g.r.forbes, s.massie, s.craw\}@rgu.ac.uk
    cclare@antillion.com
}

\begin{document}

\maketitle

\begin{abstract}

Monitoring behaviour in smart homes using sensors can offer insights into changes in the independent ability and long-term health of residents. Passive Infrared motion sensors (PIRs) are standard, however may not accurately track the full duration of movement. They also require line-of-sight to detect motion which can restrict performance and ensures they must be visible to residents. Channel State Information (CSI) is a low cost, unintrusive form of radio sensing which can monitor movement but also offers opportunities to generate rich data. We have developed a novel, self-calibrating motion detection system which uses CSI data collected and processed on a stock Raspberry Pi 4. This system exploits the correlation between CSI frames, on which we perform variance analysis using our algorithm to accurately measure the full period of a resident's movement. We demonstrate the effectiveness of this approach in several real-world environments. Experiments conducted demonstrate that activity start and end time can be accurately detected for motion examples of different intensities at different locations. 
% Both the implementation software and the collected data have been made publicly available.

\end{abstract}

\section{Introduction}

    With most countries facing an aging population, the need for solutions to support independent living at home has increased over the last decade \cite{Tian2013}. By monitoring resident behaviour using sensor networks, smart homes can offer insights into health and well-being either directly to the resident or to family and carers. While the goal of behavioural monitoring systems can vary depending on the desired scope, an ideal system is capable of performing fine-grained activity recognition. A major constituent of activity recognition is motion detection. This is important in long-term health monitoring as periods of active movement can be traced over time to indicate trends in overall activity levels.
    
    Various sensor technologies can be employed to monitor movement. The choice of technology has a direct impact on the quality of behavioural monitoring, through the richness of the data they collect, their cost, and their perceived intrusiveness to residents. Many homes already contain Passive Infrared motion sensors (PIRs), however these are event-focused and so produce small amounts of data which is insufficient for high quality behavioural insights. Alternative modalities such as wearables and cameras capture richer data with diverse applications, however residents may reject them as they are considered too intrusive for a home environment \cite{Sucerquia2017}. Radio Frequency (RF)-based solutions, such as proposed by~\cite{Adib2013}, are unintrusive but can be too expensive to deploy at a large scale. Selecting a sensor for smart housing currently requires balancing the inherent characteristics of the modality. The ideal solution would be an ambient low cost sensor which could collect rich data for high quality behavioural insights.
    
    Channel State Information (CSI) is an unintrusive form of RF-based sensing which can be performed using a wide range of hardware. In network design and optimisation, CSI is used by engineers to identify problems with wireless links and their interaction with the environment. However, in the last decade CSI has seen research applications for wider use, such as respiration and heart rate monitoring \cite{Wang2017,Khamis2018}, and gesture recognition \cite{Bu2020}. While enterprise software-defined radios (SDRs) can be used for CSI collection, capable low cost solutions are now becoming more accessible. Recent developments have allowed for CSI to be collected using the Raspberry Pi 4, one of the world's most popular embedded boards \cite{Gringoli2019}. This provides a scalable solution for CSI-based applications on a low-cost device with on-device processing capabilities. 
    
    While PIRs are effective at detecting if motion has occurred in a given environment, they do not accurately measure the duration and intensity of the motion taking place. Once activated, PIRs have a reset time of typically a few seconds, which makes accurate measurement of the extent of movement taking place difficult. Most PIRs require line-of-sight which means they are visible to residents. As some can find PIRs relatively intrusive, this can result in residents consciously or unconsciously modifying their behaviour when near PIRs~\cite{Moretti2013}. A common example of this is residents waving at the sensors, generating additional sensor activation data. CSI-based sensors can potentially eradicate performative behaviour, as they do not require line-of-sight in order to function. The device can easily be hidden behind a surface or wall. 
    
    The key contribution of this work is establishing a scalable and unintrusive solution for motion detection in smart home environments, which can replace existing PIR sensors. We introduce a novel statistical approach for accurate coarse movement detection using CSI on the Raspberry Pi 4.
    
    We have made both the software-based implementation and dataset from our experiments publicly available \footnote{Framework: \url{https://github.com/Gi-z/CSIKit}}\footnote{Data: \url{https://github.com/Gi-z/MovementDetection-Demo}}.

\section{Related Work}

    There is increasing focus on encouraging more people to continue living independently at home. Aging residents tend to face health issues, such as chronic illnesses, conditions affecting mobility, and other long term health conditions \cite{Lee2017}. These conditions may not be noticed by the resident or healthcare professionals for many years, however they may cause small changes in behaviour which can gradually progress over time. These conditions can be identified by testing at a hospital or general practitioner, but many residents avoid medical appointments due to fear of outcome or just the inconvenience. Furthermore, residents who do present for testing may subconsciously put unrepresentative effort into their performances \cite{Ayena2016a}. If elements of these tests could be performed through continuous in-home monitoring, a representative performance could be captured many times over a long term.
    
    Many in-home behavioural monitoring studies aim to identify trends which can be attributed to early onset health conditions, such as taking longer to stand up and more frequent bathroom usage \cite{Jonhson1996}. Binary smart home sensor technologies can be well-suited to monitoring these behaviours. Ogawa~\shortcite{Ogawa2002} used a variety of early smart home sensing hardware, including PIRs, magnetic switches, and appliance-specific sensors. This network of sensors supported the monitoring of regular daily behaviours, such as number of room transitions, and length of sleep. However many of the behaviours they identified could only be tracked by adding specific sensors for one activity, highlighting limitations of binary sensing modalities.
    
    CSI can be captured using commercial off-the-shelf (COTS) WiFi hardware, such as the Intel IWL5300 NIC\footnote{Intel CSI Tool: https://dhalperi.github.io/linux-80211n-csitool/}, which demonstrates similar characteristics to those observed using RF-based sensors \cite{Halperin_csitool}. This approach to sensing using CSI can be performed with reduced RF hardware capabilities, at a massively reduced cost when compared to contemporary RF-based sensing solutions. Deployment costs of CSI sensing hardware can be similar to that of standard Z-Wave connected smart PIRs. PIRs offer limited occupancy-based monitoring capabilities and require line-of-sight to their subjects, necessitating their visible presence in the environment. Sensors using CSI can offer occupancy monitoring capabilities at reduced intrusiveness, as line-of-sight is not required. Furthermore, additional functionality may become available as capabilities in the field progress, as new health monitoring applications are explored \cite{Ma2019}.
    
    There are many approaches to processing and analysing CSI data with digital signal processing, feature extraction, machine learning, and deep learning implementations all being employed for health monitoring applications using CSI \cite{Wang2019}. Yang~\shortcite{Yang2018} used a statistical frame similarity metric for occupancy monitoring using CSI data. By treating a frame of CSI data as a shape vector, they could use the Signal Tendency Index (STI) to identify the similarity of subsequent frames. A threshold was then chosen to highlight when the overall frame to frame variation rose beyond a fixed amount, indicating that the subject had entered the room. This approach showed strong results and fed into further work in activity recognition using this data. However, use of a fixed threshold offers limited flexibility for variations in the environment. Variation in distance between the subject and sensing hardware can require manual refinement of the thresholds. In addition, the hardware employed is no longer manufactured, limiting the ability to build and deploy such a system at scale.
    
    Recently, CSI collection has become possible using the Raspberry Pi 4, making it one of the cheapest and most readily available CSI collection solutions on the market. It supports the relatively modern 802.11ac protocol, and features impressive processing capabilities given the device's size, weight, and power (SWaP). This presents an opportunity to demonstrate scalable sensing solutions using low-cost hardware.

\section{Channel State Information}

    CSI has diverse capabilities which can be used for many applications, including long term behavioural monitoring. We are interested in its potential for movement detection due to the availability of low cost hardware making the widespread use of a CSI solution commercially viable for in-home health monitoring. 

    \subsection{Preliminaries}
    
        The available frequency space for 802.11 WiFi is separated into component carrier frequencies through Orthogonal Frequency-Division Multiplexing (OFDM), by which $S$ subcarriers are used to encode and transmit data independently. As each subcarrier can serve separate data streams, CSI will be different as captured from each subcarrier. In a hardware configuration using $t$ transmit antennas and $r$ receiving antennas, this can be represented in a matrix as $CSI$ for a given packet transmission $i$.
        
        \begin{equation}
            CSI_i=
            \begin{vmatrix}
                H_{1,1} & \hdots & H_{1,r}\\
                H_{2,1} & \hdots & H_{2,r}\\
                \vdots &  & \vdots\\
                H_{t,1} & \hdots & H_{t,r}\\
            \end{vmatrix}
        \end{equation}
        
        $H_{t,r}$ for a given transmit and receive pair represents a vector containing complex pairs captured for each subcarrier. The number of subcarriers available depends on the hardware configuration used for both the transmitting and receiving device and the channel bandwidth they operate on. Popular CSI hardware such as the Intel IWL5300 can access 30 subcarriers, whereas newer devices which can support 802.11ac and 80MHz channels can access around 256 subcarriers. Notably, some subcarriers function as guard carriers (or guard bands) to reduce interference, and so these remain empty by design. If the number of available subcarriers is $S$, a given $H_{t,r}$ pair can be expressed as:
        
        \begin{equation}
            H_{t,r}=\left[ h_{t,r,1}, h_{t,r,2},\hdots,h_{t,r,S} \right]
        \end{equation}
        
        The complex number $h_s$ generated for each subcarrier contains the effect of transmission on the signal from a subcarrier, from which phase $\theta$ and gain $|h_s|$ can be derived. 
        
        \begin{equation}
            h_s=|h_s|\exp{(j\cdot\theta)}
        \end{equation}
        
        Multi-path propagation is an effect inherent to wireless transmission systems, as transmitted signals do not travel directly to the receiving antenna. The combination of these external factors make up the effect of multi-path propagation:
        \begin{itemize}
            \item Reflection - Change in phase as the signal rebounds;
            \item Scattering - Variations in path affecting signal shape;
            \item Attenuation - Reduction in observed amplitude.
        \end{itemize}
        
        Static objects in the environment will affect multi-path propagation in a consistent manner, whereas dynamic objects such as human bodies will have a variable effect. This variation in the observed CSI data can be interpreted as impacts on the multi-paths caused by human activity. This may be through passive motion such as respiration, or active movement such as walking. While both phase and gain are modulated by this activity this paper will focus on the use of gain.
        
    \subsection{Hardware Accessibility}
    
        CSI can be collected using a wide range of wireless hardware, from low cost COTS WiFi devices, to enterprise-grade solutions such as Ettus Research\footnote{Ettus Research: https://www.ettus.com/} USRP software defined radios (SDR). What separates these options is their cost and their capabilities. Low cost CSI solutions tend to be more limited in their functional bandwidth, frequency band range, and their number of antennas and spatial streams. Enterprise SDRs offer fine-grained control over their radio hardware, while COTS hardware is often limited by restricted driver implementations. This may be interpreted as a weakness of COTS devices, but high end SDRs typically require radio engineering experience to use effectively. SDRs can produce significantly more data, which can offer increased richness, however this also requires additional processing hardware which further increases cost.
        
        The Raspberry Pi 4, now supporting CSI collection through Nexmon\footnote{Secure Mobile Networking Lab: \\                     https://github.com/seemoo-lab/nexmon\_csi}, features both relatively modern wireless hardware (802.11ac) and on-device processing capabilities at a low cost \cite{Gringoli2019}. Raspberry Pi hardware is planned to be manufactured for the foreseeable future and can be purchased in bulk. Lower-cost CSI hardware collection solutions exist, such as the ESP32\footnote{Espressif: https://www.espressif.com/en/products/socs/esp32}, however these devices lack sufficient on-device processing capabilities for real-time processing at the fidelity we are aiming to achieve \cite{Atif2020}. This places the Raspberry Pi 4 as an effective solution for smart home health monitoring using CSI.

\section{Movement Detection}

    Our system aims to perform movement detection using variation in the correlation of sequential CSI frames which is impacted by human movement. Periods of increased variation can be identified as being caused by the impact of the subject's body movements in the RF environment. By measuring this and applying our statistical approach, we can effectively perform both movement and occupancy detection. 
    
    This system is designed to work both as a replacement for a PIR, but could also be used to enhance their functionality by working alongside them. A functioning implementation of this system can be built at comparable cost to an off-the-shelf Z-wave connected PIR sensor, but with increased temporal precision and the ability to potentially offer additional health monitoring functionality in the future.
    
    While a raw threshold-based approach has been employed in other work, they are static approaches and do not proactively adapt to changing environmental conditions. The system proposed in this work adapts to changes in the environment through an automated calibration process. Threshold-based approaches are limited in sensitivity, which can cause movements performed further away from the sensor to produce signals of lower magnitude. We aim to mitigate this by using a proportional threshold which offers fine control over sensitivity. Furthermore, we attempted to ensure that data windows beginning with movement could also be correctly classified, as a typical statistical approach may require initial data to establish a norm.
    
    \begin{figure}[ht]
        \centering
        \includegraphics[width=\linewidth]{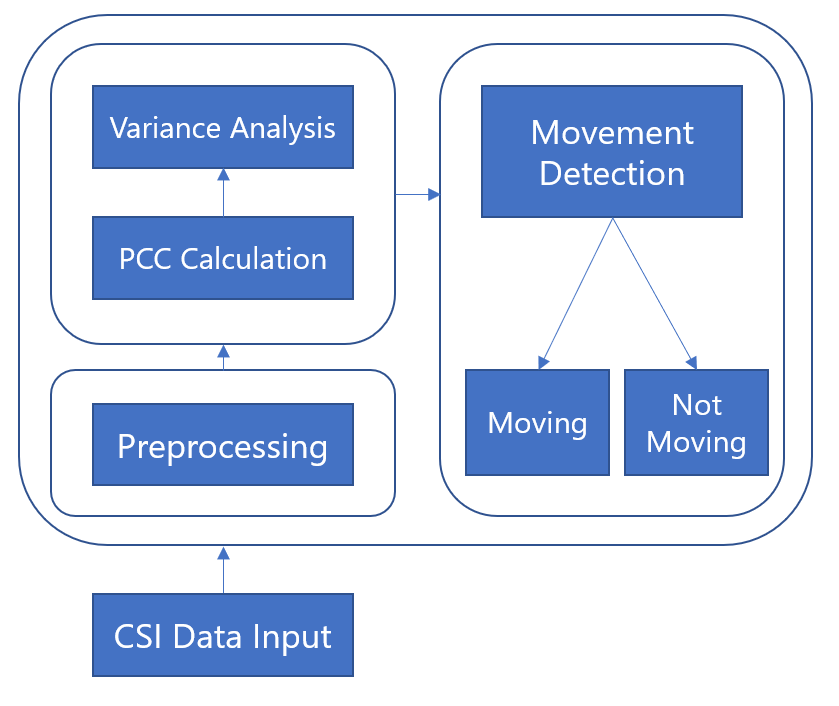}
        \caption{System overview diagram.}
        \label{fig:overview}
    \end{figure}

    The system can be split into 6 sections as seen in Figure~\ref{fig:overview}. Data is captured using CSI hardware, which can either be passed to the system as a batch, or buffered on a realtime system. The data is then preprocessed to remove noise and inconsistencies, after which the Pearson Correlation Coefficient (PCC) is calculated. Statistical analysis of variance is then performed on the resultant PCC, through which movement can be identified from the input signal. The movement output can then be relayed to the viewer as a binary value with relatively high temporal precision, as opposed to a PIR.
    
    \subsection{Preprocessing}
        
        Using the hardware configuration defined in Section 5.3, CSI amplitude data is captured at approximately 100Hz. This inconsistency is due to the ambient packet loss which can occur in a managed network solution in typical environmental conditions. The data is then linearly resampled to 100Hz. Pilot, guard, and null subcarriers are removed from the signal, from which the lower 20MHz band is extracted. A low pass filter is then applied with a lower frequency bound of 10Hz, to reduce noise from unwanted frequency components in the signal. This output is then downsampled to 10Hz to reduce processing overhead, as negligible accuracy improvements were observed using 100Hz data.
    
    \subsection{Pearson Correlation Coefficient}
        
        \begin{figure}[h]
            \centering
            \includegraphics[width=\linewidth]{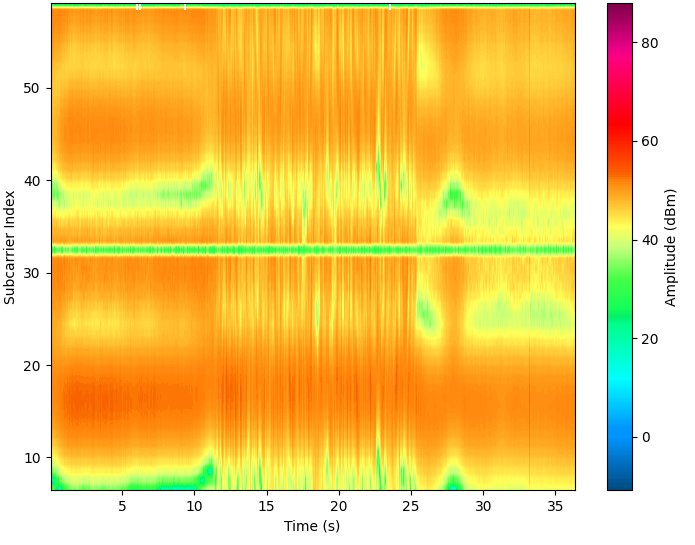}
            \caption{CSI data example showing a period of movement between 11s and 25s.}
            \label{fig:capturediagram}
        \end{figure}
    
        Human movement causes large, rapid changes in CSI, as visible in Figure~\ref{fig:capturediagram}. As a result the correlation between adjacent frames of CSI fluctuates. These fluctuations may not always be visible in a heatmap plot, however the underlying pattern in PCC variation tends to be present. We quantify the correlation between the CSI at frames $t$ and $t+1$ through the Pearson Correlation Coefficient.
        
        \begin{equation}
            PCC_{t,t+1} = \frac{cov(CSI_t,CSI_{t+1})}{\sigma (CSI_t) \cdot\sigma (CSI_{t+1})}
        \end{equation}
        
        Where $\sigma$ is the square root of the frame's variance and $cov$ is the covariance of the two frames. As the PCC is invariant under linear transformations to the data, it allows us to compare the shape of the graphs numerically. It also ensures the approach is unaffected by Automatic Gain Control (AGC), which typically negatively impacts CSI sensing performance in COTS systems.
        Other authors have used the Signal Tendency Index (STI) to quantify the changes to the CSI over time and detect movement using a threshold \cite{Yang2018}. STI is a transformation of the PCC between adjacent frames, and our testing found no benefit to using this over the PCC.
        
        \begin{equation}
            STI = \sqrt{2n(1-PCC)}
        \end{equation}
        
\begin{figure}[ht]
            \centering
            \includegraphics[width=\linewidth]{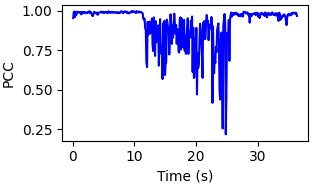}
            \caption{Pearson Correlation Coefficient for Figure~\ref{fig:capturediagram}}
            \label{fig:pcc}
        \end{figure}

        As we are using the variance of the PCC for movement detection we do not rely on its robustness, but in future work we could consider investigating Fisher's Z-Transform.
        As the extent to which PCC varies can differ significantly in real-life motion detection scenarios, we measure the variance of the PCC within two sequential 0.5 second windows, rather than identifying when the PCC has dropped below a fixed threshold. This approach ensures relative reactivity to changes in the location of the movement source, which may be closer or further away from the receiver. For example, given the PCC curve in Figure~\ref{fig:pcc}, one could reasonably assume a threshold could be established for 0.50. However, the same intensity of movement performed at the other side of the room would yield a smaller change in PCC and so not be detected. By measuring the variance between two sequential 0.5 second windows, we can monitor for a consistent relative intensity of movement, as opposed to a fixed value.
        
    \subsection{Variance Analysis}
    
        Our sliding variance analysis algorithm is designed to handle automated signal shape detection on the PCC output. It aims to overcome issues faced when using a threshold-only approach, such as inflexibility to magnitude changes. By using the running variance of the PCC as our input, spikes and continued upswings can be more robustly classified. Once calibrated, this algorithm can be used to identify periods of movement in varying environments and deployment configurations.
        
        Calibration is needed to ensure the system is capable of identifying whether an input stream contains any movement activity whatsoever. Five data captures are performed for ten seconds each in order to establish normal CSI trends in an empty room. PCC values for this data are then calculated, from which the mean and mean range are extracted. A ``containsmovement\_threshold'' can then be established as the mean minus twice the mean range. The ``mov\_threshold'' is set to 0.15, and the ``nomov\_threshold'' is set to 0.05. The ``windowsize'' is set to 5.
    
        \begin{algorithm}[tb]
            \caption{Sliding Variance Analysis}
            \label{alg:algorithm}
            \textbf{Input}: Running Variance of PCC, thresholds for movement and non-movement\\
            \textbf{Parameter}: Window size \\
            \textbf{Output}: Array of detected movement\\
            \begin{algorithmic}[1] %[1] enables line numbers
                \WHILE{$i<$length(input)}
                    \STATE X = input[i : i$+$windowsize]\\
                    \STATE Y = input[i$+$windowsize : i$+$(windowsize$*$2)]\\
                    \STATE diff = abs(mean(X)$-$mean(Y))\\
                    \IF {moving}
                        \STATE output[i] = bool(diff $>$ mov\_thresh)\\
                    \ELSE
                        \STATE output[i] = bool(diff $<$ nomov\_thresh)\\
                    \ENDIF\\
                    \STATE moving = output[i]
                \ENDWHILE\\
            \end{algorithmic}
        \end{algorithm}
        
        The ``containsmovement\_threshold'', established to indicate the presence of movement within a data capture, is used to indicate whether variance analysis should be performed. In which case a running variance filter with a window size of 10 samples is applied to the PCC and passed to Algorithm~\ref{alg:algorithm}. Iterating over each sample in the data, two sequential half second windows are established. The absolute difference between the means of both windows is then calculated. Movement can be identified as periods where the difference between the means of both windows exceeds 15\% (based on ``mov\_threshold'') of the max observed value in a given data capture. This proportional approach helps ensure that differences in magnitude of the CSI signal do not have a direct impact on the detection accuracy. Additionally, the use of a separate ``nomov\_threshold'' ensures that consistently lower variance will indicate movement has terminated, as the beginning of movement produces more erratic signals. A binary stream of movement status for each frame is then returned.

\section{Experiment}

    The objective of this experiment is to demonstrate that the system described in Section 4 can offer movement detection functionality, at least analogous to that of a PIR. CSI and PIR sensors are placed in two environments and data is collected while the resident performs a set of 5 movement-related activities. The output from both sensors is compared to the ground truth timings and a ratio of correctly-classified to incorrectly-classified time is established to score the accuracy of each sensor. Each movement type is repeated 5 times and the average result is used.
    While the CSI system may be capable of performing movement detection in through-the-wall scenarios, this experiment will establish the capabilities for movement detection within one room. With the chosen design, combined movement and presence detection can be achieved using a Raspberry Pi 4.

    \subsection{Environments}

        Data was captured from two different environments, in order to establish the system can be calibrated to environments with different external factors. These factors may include but are not limited to: interference from other WiFi devices, thickness of walls, and building density. While the impact of the environment and external factors has been identified, it has not yet been independently quantified \cite{Lee2019}. This experiment will begin to identify how performance compares between different environments, however additional experiments will be needed to more definitively demonstrate that the system is robust to environmental changes.
        
        Environment 1 is a small, open plan apartment with CSI collection performed in a 4.2x3m living room. Environment 2 is a two storey house with CSI collection performed in a home office, approximately 3x2.2m. Object clutter is representative in both environments. The ``containsmovement\_threshold'' is calculated through calibration as 0.87 and 0.92 in environments 1 and 2, respectively.
    
    \subsection{Movement Design}
    
        Five movement patterns were chosen for data collection, as listed below. These were designed to test limitations in the movement detection capabilities of both the PIR and CSI systems, while also acting as representative samples of behaviour which could be observed in smart home health monitoring scenarios. Notably, movements 4 and 5 were chosen to test the ability of the solution to respond to changes in distance between the subject and sensor while retaining similar sensitivity to both movements. 
        
        \begin{enumerate}
            \item 10 seconds of standing still, followed by 15 seconds of waving arms, and 10 seconds of standing still.
            \item 7 seconds of waving arms, followed by 7 seconds of standing still, and 7 seconds of waving arms.
            \item 10 seconds of standing still, following 10 seconds of walking, and 10 seconds of standing still.
            \item With the subject close to the sensor, 5 seconds of sitting, followed by stand up activity over 5 seconds.
            \item With the subject further from the sensor, 5 seconds of sitting, followed by stand up activity over 5 seconds.
        \end{enumerate}
        
        The subject initiated and concluded captures using a smartphone. The ground truth, against which sensor data was compared, was generated from these timelines. At 10 samples per second, accuracy is measured as the ratio of correct to incorrectly classified samples.
       
    \subsection{System Configuration}
    
        \begin{figure}[h]
            \centering
            \includegraphics[width=0.8\linewidth]{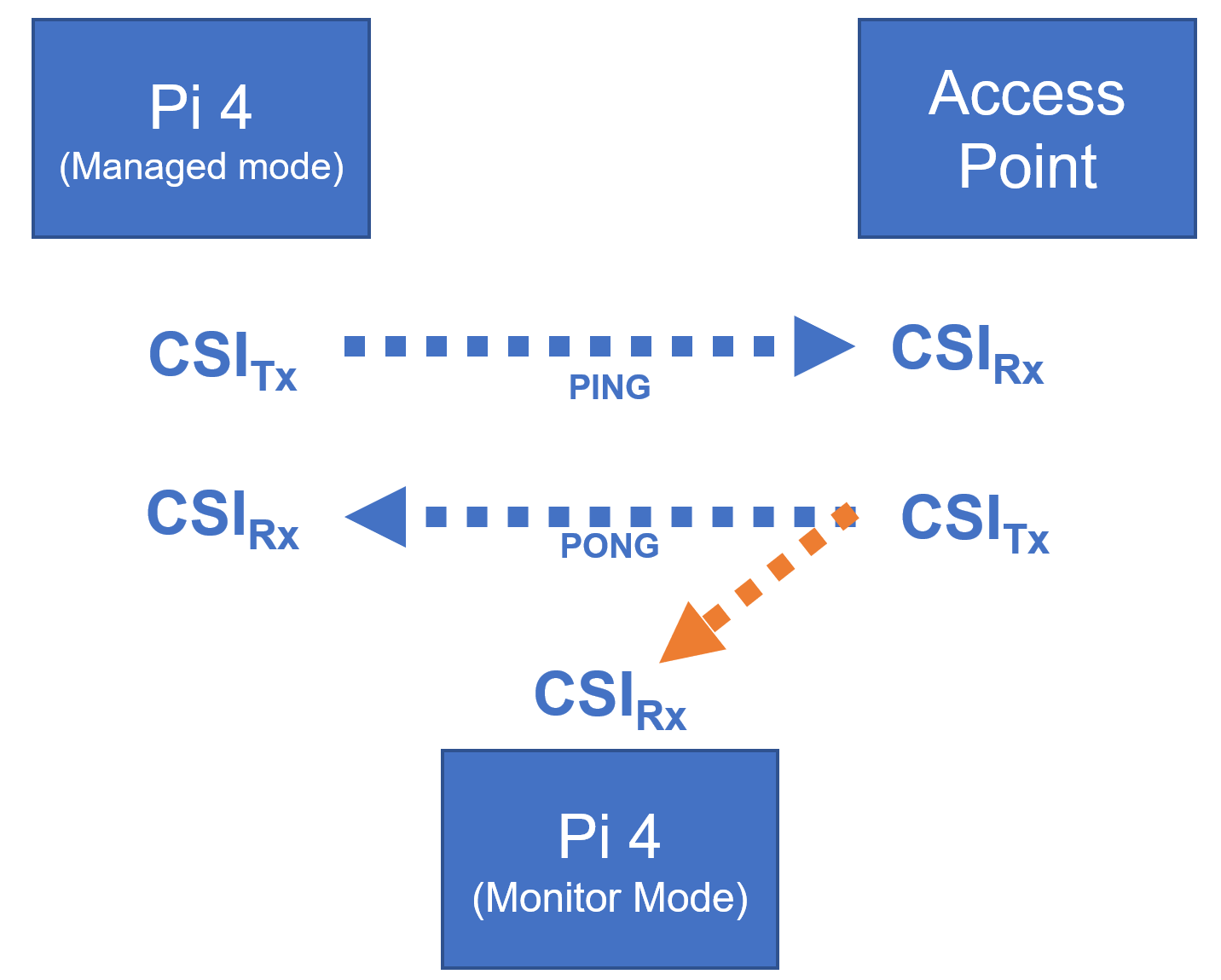}
            \caption{CSI capture device configuration.}
            \label{fig:device_layout}
        \end{figure}
    
        The collection devices are configured as shown in Figure~\ref{fig:device_layout}:
        \begin{itemize}
            \item Access Point: Sky ER110 Router running a 5GHz wireless network on channel 36 at 80MHz bandwidth;
            \item Traffic Generator: Raspberry Pi 4 running in Managed mode, sending 100Hz ping packets to the AP;
            \item CSI Collector: Raspberry Pi 4 running in Monitor mode, listening to the Traffic Generator. In addition, a BIS0001-based PIR was connected to the collector device for comparison.
        \end{itemize}
            
        This system produces CSI data at a rate of roughly 100Hz, which can then be resampled to meet the desired 100Hz source rate.
    
\section{Results}
    
    \begin{figure}[h]
        \centering
        \includegraphics[width=0.7\linewidth]{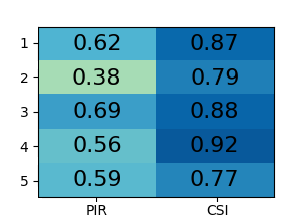}
        \caption{\textbf{Accuracy} in \textbf{Environment 1}}
        \label{fig:accuracy_env1}
        \includegraphics[width=0.7\linewidth]{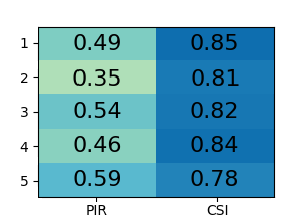}
        \caption{\textbf{Accuracy} in \textbf{Environment 2}}
        \label{fig:accuracy_env2}
    \end{figure}
    
    The performance observed shows the PIR system detects the extent of movement with an average accuracy of 56.7\% and 48.6\% in environments 1 and 2 respectively, while the CSI system does so with an average accuracy of 84.6\% and 82\%. In all movements across both environments the CSI system performs better than PIR by an average of 32\%, with a best case improvement of 46\%.
   
    \subsection{Discussion}
    
        The goal of this experiment was to identify whether our CSI movement sensor can offer similar or improved performance to a PIR. The strong performance observed indicates that our CSI sensor can outperform PIRs in monitoring the instance and duration of movement events in a smart home environment. While the activity performances captured for each movement pattern may not perfectly match up with the fixed-time ground truth, it is clear the CSI sensor is performing well with respect to both the PIR and the ground truth.
        
        Performance is largely similar between the two environments, showing the system can be accurately calibrated for different environments as described in Section 4.3. Slightly lower performance can be observed in environment 2. This may be caused by external factors affecting the RF conditions in the environment. The subject's performances for each data capture may also display poorer adherence to the ground truth timelines. Due to the increased time precision of the CSI sensor, it may be necessary to use a more sophisticated ground truth for movement capture such as video.
        
\begin{figure}[h]
            \centering
            \includegraphics[width=\linewidth]{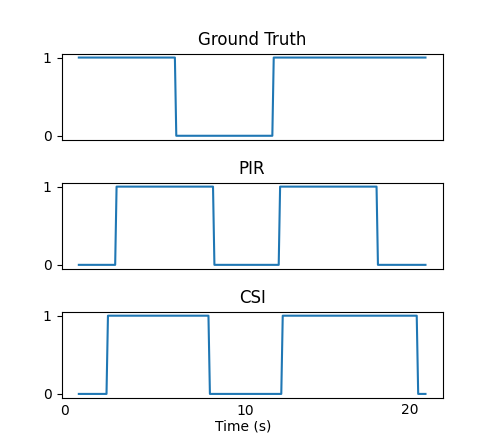}
            \caption{Binary movement output for a movement 2 capture in environment 2.}
            \label{fig:demo}
        \end{figure}

        Consistency can be observed among most of the movement styles, however there is a notable drop in performance on movement 2 samples for the PIR. Movement 2 begins and ends with movement activity, a design decision aimed to test whether the variance analysis algorithm could correctly classify samples beginning with the subject already moving. Our results showed our systems could take up to 1s to identify movement from the start of capture in these examples. As the subject controlled data capture, movement may have also in fact concluded for a short period before the data capture ended. Both the PIR and CSI sensors demonstrate this behaviour as shown in Figure~\ref{fig:demo}. However the CSI system more accurately captures both sustained movement and the termination of movement as the subject ends the capture.
        
\section{Conclusion}

    In this work we investigate the capabilities of employing CSI movement detection solutions embedded in low cost hardware for smart health applications. A novel calibration and thresholding approach has been presented which allows thresholds to be set dynamically and can be applied to CSI data to identify onset and extent of movement taking place in homes more accurately than is achieved with PIRs. 
    
    The approach has been implemented using low cost Raspberry Pi 4 hardware, and experiments demonstrate the approach to be effective by accurately capturing activity from the ground truth data. The software has been made available open source as part of a CSI framework.
    
    We believe there is an opportunity for CSI-based sensors to replace PIRs due to the increasing availability of low cost hardware. CSI sensors can be completely hidden from view and may be used to offer additional health monitoring functionality in the future. 
    
    Future developments for this system may focus on identifying instances of multiple occupancy, to allow personalised behavioural monitoring to be paused, and data fusion approaches to combine data from existing sensor hardware in smart homes with CSI sensing to enhance health monitoring capabilities at low cost.

\subsubsection{Acknowledgements}
This work was funded by Albyn Housing Society, and the Scottish Funding Council via The Data Lab innovation centre.

\bibliographystyle{named.bst}
\bibliography{library.bib}

\begin{thebibliography}{}

\bibitem[\protect\citeauthoryear{Adib and Katabi}{2013}]{Adib2013}
Fadel Adib and Dina Katabi.
\newblock See through walls with {WiFi}!
\newblock {\em ACM SIGCOMM Computer Communication Review}, 43(4):75--86, 2013.

\bibitem[\protect\citeauthoryear{Atif \bgroup \em et al.\egroup
  }{2020}]{Atif2020}
Muhammad Atif, Shapna Muralidharan, Byounghyun Yoo, and Heedong Ko.
\newblock {Wi-ESP—A tool for CSI-based Device-Free Wi-Fi Sensing (DFWS)}.
\newblock {\em Journal of Computational Design and Engineering}, 7(0):1--13,
  2020.

\bibitem[\protect\citeauthoryear{Ayena \bgroup \em et al.\egroup
  }{2016}]{Ayena2016a}
Johannes~C. Ayena, Helmi Zaibi, Martin~J.D. Otis, and Bob Antoine~J. Menelas.
\newblock {Home-Based Risk of Falling Assessment Test Using a Closed-Loop
  Balance Model}.
\newblock {\em IEEE transactions on neural systems and rehabilitation
  engineering}, 24(12):1351--1362, 2016.

\bibitem[\protect\citeauthoryear{Bu \bgroup \em et al.\egroup }{2020}]{Bu2020}
Qirong Bu, Gang Yang, Xingxia Ming, Tuo Zhang, Jun Feng, and Jing Zhang.
\newblock Deep transfer learning for gesture recognition with {WiFi} signals.
\newblock {\em Personal and Ubiquitous Computing}, 2020.

\bibitem[\protect\citeauthoryear{Gringoli \bgroup \em et al.\egroup
  }{2019}]{Gringoli2019}
Francesco Gringoli, Matthias Schulz, Jakob Link, and Matthias Hollick.
\newblock Free your {CSI}: A channel state information extraction platform for
  modern {Wi-Fi} chipsets.
\newblock In {\em Proc.s of the 13th Intl. Workshop on Wireless Network
  Testbeds, Experimental Evaluation and Characterization}, WiNTECH ’19, page
  21–28, 2019.

\bibitem[\protect\citeauthoryear{Halperin \bgroup \em et al.\egroup
  }{2011}]{Halperin_csitool}
Daniel Halperin, Wenjun Hu, Anmol Sheth, and David Wetherall.
\newblock Tool release: Gathering 802.11n traces with channel state
  information.
\newblock {\em ACM SIGCOMM CCR}, 41(1):53, Jan. 2011.

\bibitem[\protect\citeauthoryear{Johnson and Andrews}{1996}]{Jonhson1996}
P~Johnson and DC~Andrews.
\newblock Remote continuous physiological monitoring in the home.
\newblock {\em Journal of telemedicine and telecare}, 2(2):107--113, 1996.

\bibitem[\protect\citeauthoryear{Khamis \bgroup \em et al.\egroup
  }{2018}]{Khamis2018}
Abdelwahed Khamis, Chun~Tung Chou, Branislav Kusy, and Wen Hu.
\newblock Cardiofi: Enabling heart rate monitoring on unmodified {COTS WiFi}
  devices.
\newblock {\em ACM Intl. Conference Proc. Series}, pages 97--106, 2018.

\bibitem[\protect\citeauthoryear{Lee \bgroup \em et al.\egroup
  }{2017}]{Lee2017}
Gerry Lee, Nicola Pickstone, Jose Facultad, and Karen Titchener.
\newblock The future of community nursing: Hospital in the home.
\newblock {\em British journal of community nursing}, 22(4):174--180, 2017.

\bibitem[\protect\citeauthoryear{Lee \bgroup \em et al.\egroup
  }{2019}]{Lee2019}
Hoonyong Lee, Changbum~R. Ahn, Nakjung Choi, Toseung Kim, and Hyunsoo Lee.
\newblock The effects of housing environments on the performance of
  activity-recognition systems using wi-fi channel state information: An
  exploratory study.
\newblock {\em Sensors (Switzerland)}, 19(5), 2019.

\bibitem[\protect\citeauthoryear{Ma \bgroup \em et al.\egroup }{2019}]{Ma2019}
Yongsen Ma, Gang Zhou, and Shuangquan Wang.
\newblock {WiFi sensing with channel state information: A survey}.
\newblock {\em ACM Computing Surveys}, 52(3), 2019.

\bibitem[\protect\citeauthoryear{Moretti \bgroup \em et al.\egroup
  }{2013}]{Moretti2013}
Giovanni Moretti, Stephen Marsland, Debraj Basu, Gourab Sen~Gupta, MUSE Group,
  et~al.
\newblock Towards a monitoring smart home for the elderly: One experience in
  retrofitting a sensor network into an existing home.
\newblock {\em Journal of Ambient Intelligence and Smart Environments},
  5(6):639--656, 2013.

\bibitem[\protect\citeauthoryear{Ogawa \bgroup \em et al.\egroup
  }{2002}]{Ogawa2002}
M.~Ogawa, R.~Suzuki, S.~Otake, T.~Izutsu, T.~Iwaya, and T.~Togawa.
\newblock {Long term remote behavioral monitoring of elderly by using sensors
  installed in ordinary houses}.
\newblock {\em 2nd Annual Intl. IEEE-EMBS Special Topic Conference on
  Microtechnologies in Medicine and Biology - Proc.s}, pages 322--325, 2002.

\bibitem[\protect\citeauthoryear{Sucerquia \bgroup \em et al.\egroup
  }{2017}]{Sucerquia2017}
Angela Sucerquia, Jos{\'{e}}~David L{\'{o}}pez, and Jes{\'{u}}s~Francisco
  Vargas-Bonilla.
\newblock {SisFall: A fall and movement dataset}.
\newblock {\em Sensors (Switzerland)}, 17(1), 2017.

\bibitem[\protect\citeauthoryear{Tian \bgroup \em et al.\egroup
  }{2013}]{Tian2013}
Y~Tian, James Thompson, D~Buck, and L~Sonola.
\newblock Exploring the system-wide costs of falls in older people in {Torbay}.
\newblock {\em London: The King's Fund}, pages 1--12, 2013.

\bibitem[\protect\citeauthoryear{Wang \bgroup \em et al.\egroup
  }{2017}]{Wang2017}
Xuyu Wang, Chao Yang, and Shiwen Mao.
\newblock {PhaseBeat: Exploiting CSI Phase Data for Vital Sign Monitoring with
  Commodity WiFi Devices}.
\newblock {\em Proc.s - Intl. Conference on Distributed Computing Systems},
  pages 1230--1239, 2017.

\bibitem[\protect\citeauthoryear{Wang \bgroup \em et al.\egroup
  }{2019}]{Wang2019}
Zhengjie Wang, Kangkang Jiang, Yushan Hou, Zehua Huang, Wenwen Dou, Chengming
  Zhang, and Yinjing Guo.
\newblock {A Survey on CSI-Based Human Behavior Recognition in Through-The-Wall
  Scenario}, 2019.

\bibitem[\protect\citeauthoryear{Yang \bgroup \em et al.\egroup
  }{2018}]{Yang2018}
Jianfei Yang, Han Zou, Hao Jiang, and Lihua Xie.
\newblock {Device-Free Occupant Activity Sensing Using WiFi-Enabled IoT Devices
  for Smart Homes}.
\newblock {\em IEEE Internet of Things Journal}, 5(5):3991--4002, 2018.

\end{thebibliography}

\end{document}